\documentclass[a4paper]{article}

\usepackage{INTERSPEECH2022}
\usepackage{enumitem}

\usepackage{cuted}
\usepackage{graphicx}
\usepackage{booktabs}
\usepackage{multirow}
\usepackage[hyphens]{url}

\usepackage{hyperref}

\usepackage{color}
\usepackage[normalem]{ulem}

\title{Representing `how you say' with `what you say': English corpus of focused speech and text reflecting corresponding implications}
\name{Naoaki Suzuki, Satoshi Nakamura}
\address{
  Nara Institute of Science and Technology, Japan}
\email{\{suzuki.naoaki.sg4, s-nakamura\}@is.naist.jp}

\begin{document}

\maketitle
\begin{abstract}
    In speech communication, how something is said (paralinguistic information) is as crucial as what is said (linguistic information). As a type of paralinguistic information, English speech uses sentence stress, the heaviest prominence within a sentence, to convey emphasis. While different placements of sentence stress communicate different emphatic implications, current speech translation systems return the same translations if the utterances are linguistically identical, losing paralinguistic information. Concentrating on focus, a type of emphasis, we propose mapping paralinguistic information into the linguistic domain within the source language using lexical and grammatical devices. This method enables us to translate the paraphrased text representations instead of the transcription of the original speech and obtain translations that preserve paralinguistic information. As a first step, we present the collection of an English corpus containing speech that differed in the placement of focus along with the corresponding text, which was designed to reflect the implied meaning of the speech. Also, analyses of our corpus demonstrated that mapping of focus from the paralinguistic domain into the linguistic domain involved various lexical and grammatical methods. The data and insights from our analysis will further advance research into paralinguistic translation. The corpus will be published via LDC and our website \footnote{\url{https://dsc-nlp.naist.jp/data/speech/paralinguistic_paraphrase/}}.
  
\end{abstract}

\section{Introduction}
    In speech communication people make use of two types of information to convey their intentions: \textit{what} is said (linguistic information) and \textit{how} it is said (paralinguistic information) \cite{mitra2019leveraging}. Paralinguistic information is expressed by suprasegmental features such as duration, intensity and pitch. Even with the same linguistic information, changes in these prosodic features can communicate different implications. One such implication involves emphasis. As the terminology can be ambiguous, we follow Kohler's \cite{kohler2006emphasis} distinctions of two kinds of emphasis: \textit{emphasis for focus}; which `singles out elements of discourse by making them more salient than others'; and \textit{emphasis for intensity}; which `intensifies the meaning contained in the elements.' The current work addresses the first category of emphasis: focus. In the literature of a semantic framework called Alternative Semantics \cite{rooth1985association, rooth1992theory}, focus is understood as the indication of `the presence of alternatives that are relevant for the interpretation of linguistic expressions' \cite{krifka2008basic}. Consider this example:
    
        \begin{enumerate}[label=(\arabic*)]
            \item \begin{enumerate}[label=\alph*.]
                      \item \label{emotive_a} \textbf{John} bought the apple.
                      \item \label{emotive_b} John \textbf{bought} the apple.
                  \end{enumerate}
        \end{enumerate}
        
    \noindent In (1a), \textit{John} is focused, as indicated by bold typeface. The speaker implies that \textit{It was John who bought the apple}, indicating the presence of contextually possible alternatives such as \textit{Peter} or \textit{Mary}, and at the same time indirectly ruling out these agents for the person who \textit{bought the apple}. In contrast, in (1b), focus falls on \textit{bought}, by which the speaker indicates that \textit{What John did was not sell the apple, but buy the apple}. Although (1a) and (1b) give identical linguistic information, the differences in the focused words create different connotations. In English speech, focus is marked by sentence stress, the most prominent stress within a sentence \cite{wells2006english}. To avoid misunderstanding, interlocutors must correctly perceive the placement of sentence stress and understand the associated focused meaning. While this inherent skill is taken for granted among native English speakers, non-natives find it challenging \cite{derwing2012longitudinal}. Therefore, cross-lingual interactions demand an automated system that can output the implied meaning for non-natives. \\
    \indent The idea of speech translation (ST), which automatically translates speech in a source language (SL) to text or speech in a target language (TL), might be helpful toward achieving such cross-lingual communication. In recent years, ST systems have made significant progress in translating linguistic information correctly. However, most of the ST systems developed so far have not attained the capability to consider paralinguistic information, including focus. Translation by these systems is based on the transcription produced by automatic speech recognition (ASR), which is designed to transcribe only the contents of an utterance, resulting in the loss of paralinguistic information. The current ST models translate (1a) and (1b) the same way, even though they convey different implications. Recently, however, as the importance of paralinguistic information has become more widely acknowledged, some studies have tackled the translation of paralinguistic information such as voice quality \cite{jia2019direct}, emotions \cite{aguero2006prosody, akagi2014emotional} and emphasis \cite{aguero2006prosody, kano2012method, kano2013generalizing, anumanchipalli2012intent, tsiartas2013toward, do2016transferring, do2016preserving, do2018sequence} by mapping the prosodic cues in the SL, such as intensity, duration and fundamental frequency, to speech in the TL.
    
    \begin{figure*}
        \centering
        \includegraphics[width=0.75\textwidth]{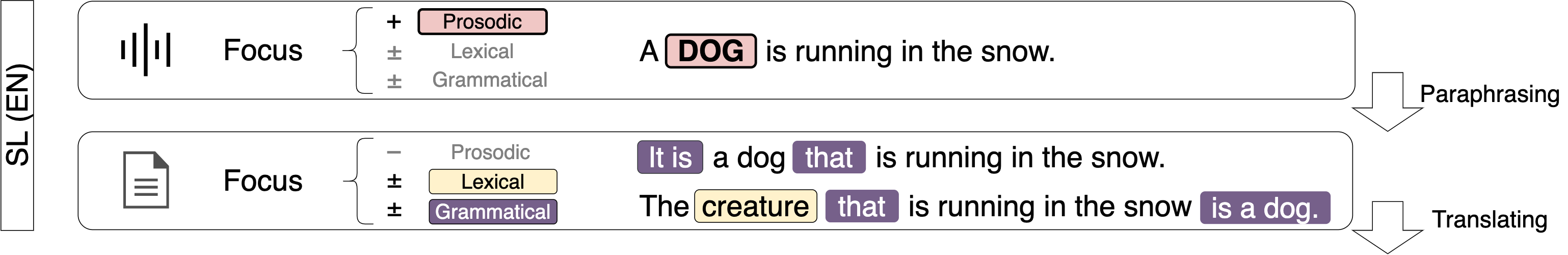}
        \caption{Example of focus translation through paraphrased text. Before translating, within the SL, focus expressed by prosodic device in the original speech is transformed into that expressed with lexical and grammatical devices in text.}
        \label{fig: proposed_method}
    \end{figure*}
    
    While such efforts have advanced the research on translation of paralinguistic cues, it cannot be assumed that the TL necessarily has a prosodic counterpart that plays the same role as that in the SL. For instance, for information focusing, languages make use of not only prosodic devices but also lexical and grammatical devices \cite{cruttenden1997intonation}. Although English speech often uses sentence stress rather than linguistic devices for focus \cite{cruttenden1997intonation}, the degree of reliance on such methods vary from language to language \cite{cruttenden1997intonation, downing2006prosody, hartmann2008focus}. This observation partially limits the efforts of an acoustic-to-acoustic way of paralinguistic translation. Instead, we argue the potential of the acoustic-to-linguistic mapping of paralinguistic information by paraphrasing the speech into the linguistic domain within the SL, with the help of lexical and grammatical devices, and then passing the paraphrases on to the translation module (Figure \ref{fig: proposed_method}). Such a method could return different translations depending on the focused items, even with the same linguistic information. 
    
    The achievement of acoustic-to-linguistic focus transformation requires data for training such a model. However, to the best of our knowledge, no existing English corpus contains pairs of: speech having different items in focus; and text reflecting the relevant implications. In the most related work, an English corpus \cite{do2018toward} was built with text items that differ in the degree of \textit{emphasis for intensity}, e.g. (\textit{It is a little bit hot} / \textit{It is extremely hot}), and the corresponding speech, e.g. (\textit{It is \textbf{hot}} / \textit{It is \textbf{HOT}}) which was recorded so that it would represent the same degree of emphasis as that in the text. Since their focus was on the degree of emphasis such as weak/strong, every speech sample fixed its position of focus on an adjective. The data was later used for building a model of speech-to-text emphasis translation \cite{tokuyama2021transcribing}. A recent work in text-to-speech (TTS) \cite{latif2021controlling} added the elements of \textit{emphasis for focus} to the previous efforts. Speech items which differed in the placement of focus were collected, e.g. (\textit{\textbf{Sarah} closed the door}, \textit{Sarah \textbf{closed} the door}). However, since the purpose of the work was to achieve prosodic prominence in a TTS model from annotated text, the data did not involve the corresponding implications of speech. Also, while words in a closed-class, e.g. \textit{the, is}, can be focused \cite{greenbaum1990student}, such cases remain to be addressed. \\
    \indent We also argue the importance of understanding the relationships of focus representations in speech and text, i.e. how focus expressed in speech is paraphrased or transformed into the linguistic domain. In the literature, some studies addressed patterns of general paraphrase alternations, i.e. how an original text item is paraphrased into another one which has approximately the same meanings \cite{boonthum2004istart, dolan2004unsupervised, vila2014paraphrase}, and others examined how linguistic items can convey focus \cite{taglicht1984message, rooth1985association}. However, it has not been clear the methods of mapping focus from the paralinguistic dimension to the linguistic one. To these ends, we present the creation of an English corpus containing speech that differs in the placement of focus, where words in a closed-class are also the targets of focus, and text reflecting the relevant implications. We also perform quantitative and qualitative analyses of the transformation of focus from the paralinguistic domain to the linguistic domain. 
    
\subsection{Focus in English}
    This section briefly summarises how the English language employs prosodic and linguistic devices to convey focus.
    \subsubsection{Prosodic device}
        As mentioned in the previous section, English speech uses sentence stress to convey focus. Native English speakers highlight certain information and draw a listener's attention using the following procedure \cite{wells2006english}. Depending on the intention, they first break spoken materials into smaller chunks called intonation phrases (IPs). Then, in each IP they select the most important word and put sentence stress on that word's stressed syllable, i.e. syllable that has lexical stress.  
        

    \subsubsection{Lexical and grammatical devices}
        Prosody is not the only device to convey focus. Lexical and grammatical devices are also available \cite{cruttenden1997intonation}. As one type of the lexical devices, a group of words called focus particles can convey focus \cite{konig2002meaning}. For instance, \textit{only}, \textit{even}, and \textit{alone} can serve this purpose \cite{rooth1985association}. Consider the following example:
        
        \begin{enumerate}[label=(\arabic*)]\addtocounter{enumi}{1}
            \item   \begin{enumerate}[label=\alph*: ]
                        \item \label{lexical_a} John bought an apple.
                        \item \label{lexical_b} John bought only an \underline{apple}.
                    \end{enumerate}
        \end{enumerate}
        
        \noindent Compared to (2a), (2b) explicitly states what \textit{John bought} was neither \textit{an orange} nor \textit{a banana}, resulting in \textit{an apple} being highlighted, as indicated by the underline. Similarly, \textit{let alone} \cite{fillmore1988regularity}, reflexive pronouns such as \textit{himself}/\textit{herself} \cite{konig2006focused}, \textit{particularly, mainly}, and many more such words \cite{konig2002meaning} can be used as lexical items for focus. \\
        \indent A grammatical device can also perform focus by changing the structure of a sentence. Grammatical items include constructions of cleft (\textit{It was \underline{Simon} who kicked the door.}), pseudo-cleft (\textit{What Mary bought was an \underline{apple}}), inversion (\textit{And then appears a \underline{bear}}), and passivization (\textit{I was bit by a \underline{dog}}) \cite{greenbaum1990student}. It should be noted that grammatical reconstructions often involve shifting the target of focus toward the end of the sentence, since English information structure is governed mainly by the principles \textit{Given-Before-New} and \textit{End Weight} \cite{aarts2011oxford}.

\section{Corpus construction}
    \begin{figure*}[t]
            \centering
            \includegraphics[width=0.90\textwidth]{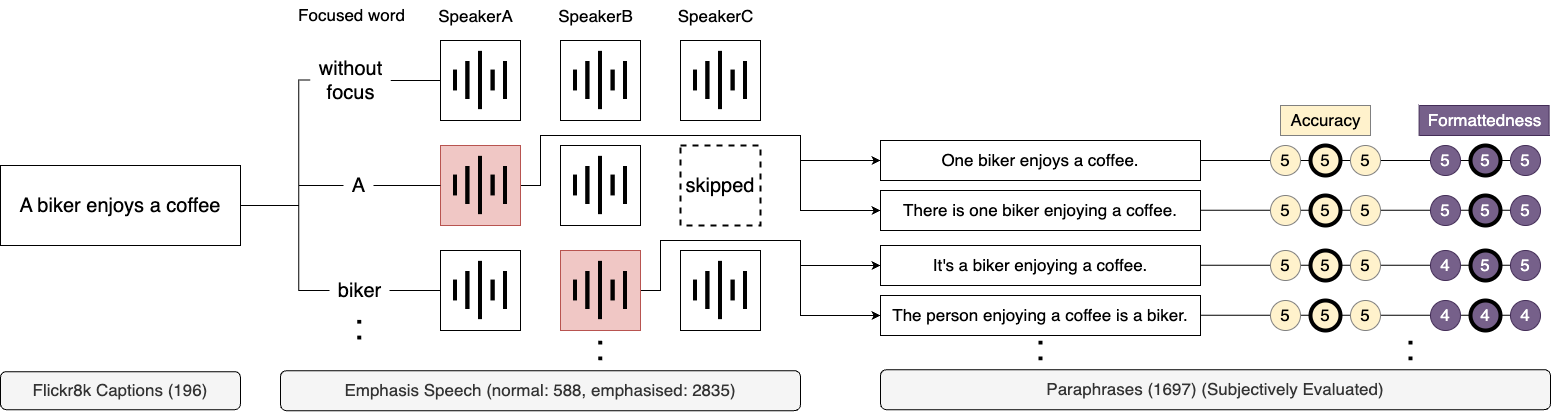}
            \caption{Corpus Design}
            \label{fig:overview}
        \end{figure*}
        
\subsection{Text design}
    We started the text construction from Flickr8k \cite{rashtchian2010collecting}, which consists of over 8000 images that depict actions relating to people or animals. Five text descriptions are given for each image, with over 40,000 annotations. To select optimal captions, we removed sentences that fell under one of the following conditions: including punctuation other than a period; corrected by GECToR \cite{omelianchuk2020gector}, a grammatical error correction model; a noun phrase; more than six words; identical to another caption. We left only one caption if an image had multiple captions. Regarding the sentence length, for simplicity, we wanted each speech to have only one sentence stress, meaning that an entire sentence is treated as a single IP. To determine an optimal sentence length for a single IP, we examined the London-Lund corpus of spoken English \cite{svartvik1990london}, which consists of half a million words and prosodic transcriptions, including tone units, or IPs. We calculated word length for each IP, and chose 6, which equaled the third quartile in the corpus, as the maximum sentence length. After the filtering, we had 1375 sentences and selected the first 196 captions as the source sentences for our corpus. The text items included \textit{A biker enjoys a coffee}, for instance (Figure \ref{fig:overview}).
    
\subsection{Speech collection}
    We employed Amazon's Mechanical Turk (MTurk) for data collection, a crowdsourcing platform allowing researchers to create tasks called HITs (Human Intelligence Task) and anonymous users (Workers) to complete them for a small monetary fee.
    
    \subsubsection{Recording application}
    MTurk does not provide an interface for audio collection, so we created a web-based recording application that could capture a user's recordings interactively and store them in a back-end server. To encourage subjects to speak properly, we enabled Google speech-to-text API so that the system could give the instant feedback `Speak Clearly' if it did not recognise incoming speech as English words. Speakers could use basic functions such as start and stop of the recordings, and re-recording. As an additional feature, for each utterance, we recorded an extra two seconds at the end while the speech evaluation was in progress; this was done to capture the environment's sound, and during this time, users were instructed to remain silent.
    
    \subsubsection{Recording procedure}
    We attempted to collect a set of speech samples, each with different placement of focus, for each caption using three different speakers (Figure \ref{fig:overview}). For future use, normal readings without focus was also collected from the same speakers. We asked Workers located in the UK to participate in the tasks. In the recording HITs, Workers were instructed as follows: Look at the displayed written sentence, e.g. (\textit{Two \underline{men} are ice fishing}); Make a recording emphasising the underlined word if it does not sound unnatural, otherwise, skip this recording; If there is no underlined word, e.g. (\textit{Two men are ice fishing}), read the sentence in a normal way. \\
    \indent When using a crowdsourcing service, taking measures to ensure the quality of data is crucial \cite{rashtchian2010collecting, kennedy2020shape}. We prepared a qualification test, in which Workers needed to fill out general information such as age, gender and accent after an agreement. Then they conducted the step described above for three captions. After manually checking the results, we allowed those who made recordings as instructed to proceed to the main recording HITs. The collection resulted in 3423 speech items (normal: 588, emphasised: 2835) by 9 British natives (4 male, 5 female, age range 22--60 years with a median of 37 years). We paid Workers \$1.20 for processing two captions, and \$2 for three, resulting in \$30.9 per hour on average, surpassing \$15, which is considered to be fair among Workers in MTurk \cite{whiting2019fair}.

\subsection{Paraphrase collection}
\begin{figure}[h]
    \centering
    \includegraphics[width=0.71\linewidth]{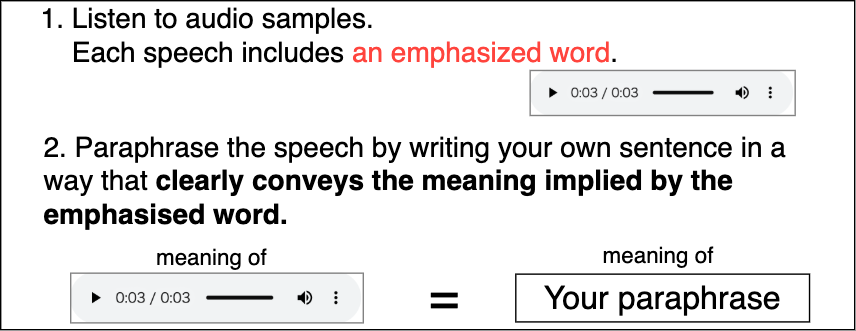}
    \caption{Instructions for paraphrase collection}
    \label{fig:paraphrase_instruction}
\end{figure}
    Free-form tasks, including the paraphrase collection, need to make sure that participants are native speakers \cite{rashtchian2010collecting}. To select native English speakers, we published tasks for Workers residing in an English speaking country (US, CA, UK, AU, NZ) and asked them what their first language was. 600 Workers participated in the task; 550 answered English as their first language. We regarded the 550 Workers as native English speakers, and divide them in half into two groups: G1 and G2.\\
    \indent Figure \ref{fig:paraphrase_instruction} shows a screenshot of the main instructions given for the paraphrase HITs. Workers were asked to listen to a set of focused speech samples for one caption, where each speech item was randomly selected from available recordings (Figure \ref{fig:overview}), and then told to come up with a written sentence for that speech item which clearly conveyed the implied meaning in the emphasised speech. We encouraged Workers to freely make changes in grammar and vocabulary but not to use capitalisation, an exclamation mark, or the addition of their own inferences or text providing situational context. We prepared a qualification test for the native speakers in G1, in which they needed to make a brief summary of the instructions in their own words as well as complete the paraphrasing tasks. After manually checking the results, we allowed those who completed the task as instructed to start the main paraphrase HITs. Two workers performed each HIT, obtaining two paraphrases for each emphasised speech sample. We paid \$0.15 for each paraphrasing, resulting in \$16.5 per hour on average. We obtained 2130 paraphrases from 16 Workers.

\subsection{Filtering paraphrases}
    The quality of the paraphrases produced by the Workers will influence future research using the corpus. To evaluate the quality, we set subjective evaluation HITs. Workers were presented with emphasised speech and a paraphrase and asked to evaluate its quality from two perspectives on a scale from one to five: (A) how accurately the paraphrase conveyed the implied meaning in speech; (B) how well-formatted the paraphrase sentence is. We set a qualification task for the native speakers in G2 in the same way as the paraphrase collection and asked those who passed to join the main evaluation HITs (16 Workers). Three different Workers evaluated each paraphrase. We calculated median values for both perspectives (Figure \ref{fig:overview}), and samples with a value of 3 or less for perspective (A) and 2 or less for (B) were discarded. We collected the paraphrases again for those removed samples, and repeated the same procedure of evaluation and removing as above. We paid \$0.10 for evaluating each paraphrase, resulting in \$25.0 per hour on average.
    
    After the evaluation in MTurk, we filtered out paraphrases which fell one of the following conditions: paraphrases with relatively high variance of the accuracy evaluation scores ($\sigma^2$ $>$ 1.6), e.g. scores: 2, 5, 5); The tense changed from present to past, e.g. (\textit{The men are climbing.} / \textit{The men were climbing on something.}) After the filtering, we had 1697 paraphrases. Table \ref{tab:corpus_example} shows an example of speech and paraphrase pairs.

    \begin{table}[h]
    \caption{Example speech-paraphrase pairs in the corpus}
    \scriptsize
    \centering
    \begin{tabular}{c|c}
        Focused speech & Paraphrase \\
        \hline
        \textbf{A} biker enjoys a coffee & One biker enjoys a coffee \\
        \textbf{A} biker enjoys a coffee & There is one biker enjoying a coffee \\
        A \textbf{biker} enjoys a coffee & It's a biker enjoying a coffee \\
        A \textbf{biker} enjoys a coffee & The person enjoying a coffee is a biker \\
        A biker \textbf{enjoys} a coffee & A biker drinking a coffee is enjoying it \\
        A biker \textbf{enjoys} a coffee & The biker seems to enjoy a coffee \\
        A biker enjoys \textbf{a} coffee & A biker enjoys one coffee \\
        A biker enjoys \textbf{a} coffee & A biker has one coffee he enjoys \\
        A biker enjoys a \textbf{coffee} & What the biker enjoys is a coffee \\
        A biker enjoys a \textbf{coffee} & It is coffee the biker enjoys \\
    \end{tabular}
    \label{tab:corpus_example}
\end{table}

\section{Analysis}
    To explore how focus in speech was mapped into the linguistic domain, we hand-examined the samples of speech-paraphrase pairs and made broad categorisations of the transformation patterns as follows; from lexical and grammatical perspectives (the patterns do not cover all the transformation methods). 
    
    
    \begin{itemize}[leftmargin=*]
        \item Lexical transformations
            \begin{itemize}[leftmargin=15pt]
                \item \label{substitution} \textbf{Substitution}:~substitute the focused word with its synonyms, e.g. (\textit{\textbf{dog}} / \textit{canine}), (\textit{\textbf{on}} / \textit{on top of}).
                
                \item \label{modification} \textbf{Modification}:~modify the focused word or its phrase with modifiers such as adverbs or a clause, e.g. (\textit{\textbf{play}} / \textit{play actively}), (\textit{\textbf{is}} / \textit{is indeed}).
                
                \item \label{negation} \textbf{Negation}:~explicitly state an alternative of the focused word and negate it (\textit{A \textbf{man}} / \textit{A man, not a woman}).
            \end{itemize}
            
        \item Grammatical transformations
            \begin{itemize}[leftmargin=12pt]
                \item \label{leftward} \textbf{Leftward shift}:~move the focused word towards the beginning of the sentence. Grammatical constructions such as cleft, reversed-pseudo-cleft and inversion were used, e.g. (\textit{\textbf{People} sit .. } / \textit{It's people who sit ..}), (\textit{.. play \textbf{baseball}} / \textit{baseball is what .. play}).
                
                \item \label{rightward} \textbf{Rightward shift}:~move the focused word towards the end of the sentence. Grammatical constructions such as pseudo-cleft, inversion and other methods were used, e.g. (\textit{Children \textbf{play} ..} / \textit{what children do .. is play}), (\textit{ .. is \textbf{rock} climbing} / \textit{.. is climbing a rock})
                
                \item \textbf{Tense change}:~change the tense from simple to progressive or vice versa, e.g. (\textit{\textbf{is} walking} / \textit{walks}).
            \end{itemize}
    \end{itemize}
    
    Furthermore, we observed that a certain part-of-speech is more likely to use a certain transformation method. To quantify this tendency, following a method taken for analysis of paraphrase alternations \cite{dolan2004unsupervised}, we randomly sampled 50 paraphrases for each part-of-speech if it had more than 50 occurrences in the corpus and compared the frequency of each transformation method. We counted each phenomenon for each sentence. We restricted the counting only if the focused word or the phrase of the focused word undergoes transformations listed above. Consider this example:  (\textit{a dog trots through \textbf{the} grass} / \textit{\underline{the only grassy area} is what a dog trots through}). In this case, not only the focused word \textit{the}, but also the noun phrase it involved, indicated by underline, was under investigation; \textit{only} was inserted (modification), and \textit{grass} was replaced with \textit{grassy area} (substitution). As a grammatical method, reversed-pseudo-cleft (leftward shift) was used. Table \ref{tab: phenometa} shows the average number of occurrences of each transformation in each focused part-of-speech. The pattern of the tense change was counted only when the focused word was verbs or auxiliary verbs.
    
    \begin{table}[h]\centering
        \caption{Mean occurrences of each transformation method per part-of-speech (N: Noun, V: Verb, Adj: Adjective, Num: Numeral, Aux: Auxiliary, P: Preposition, Det: Determiner)}\label{tab: phenometa}
        \tiny
        \begin{tabular}{llccccccc}
        \toprule
        &&N	&V & Adj &Num & Aux & P & Det \\
        \midrule
        \multirow{3}{*}{Lexical}&Substitution &0.28 & 0.12 & 0.10 & 0.18 & 0.08 & 0.42 & 0.72 \\
                                &Modification &0.00 & 0.02 & 0.12 & 0.06 & 0.60 & 0.18 & 0.50\\
                                &Negation &0.10 & 0.06 & 0.06 & 0.12 & 0.10 & 0.12 & 0.00 \\
        \midrule
        \multirow{3}{*}{Grammatical}&Leftward  &0.10 & 0.20 & 0.04 & 0.08 & 0.00 & 0.10 & 0.00 \\
                                    &Rightward &0.46 & 0.44 & 0.70 & 0.52 & 0.08 & 0.26 & 0.02\\
                                    &Tense     &-& 0.28 & -&-&0.20 & -& - \\
        \bottomrule
        \end{tabular}
        \label{tab:phenomena}
    \end{table}
    
\section{Discussion and conclusion}
    This study set out to create a corpus containing focused speech, where each speech item differs in the placement of focus, and the corresponding text which paraphrases the speech together with its paralinguistically expressed implications. Through our data analysis, we show that the transformation of focus information from the paralinguistic domain to the linguistic domain makes use of a variety of lexical and grammatical devices, and reliance on these methods vary depending on the syntactic category of the focused word. Many of the transformation methods observed in data were the types that have been reported as such devices in the literature, e.g. cleft-constructions in the leftward shift and pseudo-clefting in the leftward shift and negation, which serves the exact purpose of focus defined earlier; indicating the presence of alternatives. On the other hand, we also found some interesting methods such as lexical substitution, e.g. (\textit{\textbf{man}} / \textit{male adult}). \\
    \indent One of the limitations of the current study is lack of context; \textit{Why do I/they emphasise this word?} In regular paraphrasing tasks, paraphrases and their evaluations can vary depending on context \cite{barzilay2001extracting}. The importance of context would also be the case for the collection of focused speech and the corresponding text. In future work, we will consider presenting contextual information when collecting recordings and paraphrases. \\
    \indent Despite the limitations, the current work added a new direction toward further improvement of paralinguistic translation; we demonstrated the possibility of mapping paralinguistic information into the linguistic domain. The corpus and insights from our analysis will lead us to construct a ST model which uses the paraphrased text to preserve paralinguistic information. 
    
\section{Acknowledgement}
Part of this work is supported by JSPS KAKENHI Grant Number JP21H05054.

\bibliographystyle{IEEEtran}

\bibliography{draft_suzuki}

\end{document}